\documentclass{llncs}
\usepackage{makeidx}  
\usepackage{graphicx}
\usepackage{amsmath,amssymb} 
\begin{document}
\frontmatter          
\pagestyle{headings}  
\mainmatter              
\title{Adaptive Mesh Representation and Restoration of Biomedical Images}
\author{Ke Liu, Ming Xu, Zeyun Yu \thanks{Corresponding author: University of Wisconsin Milwaukee, Milwaukee, WI 53211, USA}}

%
\institute{University of Wisconsin Milwaukee, Milwaukee\\
\email{yuz@uwm.edu},\\ WWW home page:
\texttt{https://pantherfile.uwm.edu/yuz/www/bmv/index.html}}

\maketitle              

\begin{abstract}
The triangulation of images has become an active research area in recent years for its compressive representation and ease of image processing and visualization. However, little work has been done on how to faithfully recover image intensities from a triangulated mesh of an image, a process also known as image restoration or decoding from meshes. The existing methods such as linear interpolation, least-square interpolation, or interpolation based on radial basis functions (RBFs) work to some extent, but often yield blurred features (edges, corners, etc.). The main reason for this problem is due to the isotropically-defined Euclidean distance that is taken into consideration in these methods, without considering the anisotropicity of feature intensities in an image. Moreover, most existing methods use intensities defined at mesh nodes whose intensities are often ambiguously defined on or near image edges (or feature boundaries). In the current paper, a new method of restoring an image from its triangulation representation is proposed, by utilizing anisotropic radial basis functions (ARBFs).  This method considers not only the geometrical (Euclidean) distances but also the local feature orientations (anisotropic intensities).  Additionally, this method is based on the intensities of mesh faces instead of mesh nodes and thus provides a more robust restoration. The two strategies together guarantee excellent feature-preserving restoration of an image with arbitrary super-resolutions from its triangulation representation, as demonstrated by various experiments provided in the paper.

\end{abstract}
%
\section{Introduction}

Modern imaging technologies often digitize an image into a uniform array of pixels (or voxels in 3D). With uniformly sampling, the sampling density is inevitably too high in regions where intensities change slowly and too low in regions whose intensities change rapidly. Despite the ease of use in both hardware and software developments, uniformly-digitized images often pose challenges in data storage and transmission, as well as image processing, especially in 3D medical images that have been consistently and significantly grown in size in recent years. Evolving from previously commonly-used uniform sampling, non-uniform sampling and adaptive mesh triangulation of an image has become an active research area in image processing. Image triangulation involves partitioning an image into a collection of non-overlapping small triangles called mesh elements (faces or triangles). This procedure often serves as an image coding method, meaning that an image in pixels is compressed by using a number of ``super-pixels''. This method is a compact way to represent images for effective data storage and transmission, and also an efficient way to process and visualize images, especially for 3D images where the number of voxels can be extremely large. In addition, the resulting mesh edges are expected to be well aligned with image featured (edges or corners) in order to maintain a faithful restoration of the original image. Mesh modeling of an image has many applications like image compression \cite{aizawa95,davoine96,benoit-cattin99,demaret00}, motion tracking and compensation \cite{wang04,altunbasak97,toklu00,marquant00,nosratinia01,hsu01}, image processing by geometric manipulation \cite{garcia00}, medical image processing \cite{singh98}, feature detection \cite{coleman02}, pattern recognition \cite{petrou06}, computer vision \cite{sarkis07}, restoration \cite{brankov03}, tomographic reconstruction \cite{brankov04}, interpolation \cite{su03,su04} and image/video coding \cite{adams09,ramponi01,lechat97,wang96,hung03,adams08}.

A common procedure of image triangulation consists of two steps: 1) generating mesh nodes (vertices) by choosing a set of sampling points defined in the image domain, and 2) connecting these mesh nodes by Delaunay triangulation \cite{delaunay34}. Delaunay triangulation is a geometric operator and can avoid long and thin triangles that often lead to poor approximations. The selection of sampling nodes, however, is data-dependent, where the connectivity of the triangulation depends on the data set, based on which the mesh nodes are generated. Depending on how to generate mesh nodes, there are two categories of the image triangulation. The first one places mesh nodes inside the image features but near both sides of feature edges.  So the triangulated images of this category show double-layer vertices at both sides of feature edges.  The second category places mesh nodes directly at the feature edges, thus there are only single-layer vertices defined right on feature edges.  Yang et al. \cite{yang03} employed Floyd-Steinberg error-diffusion (ED) algorithm to place mesh nodes so that their spatial density varies according to the local image content.  As a result, the triangulated images fall into category I.  Adams \cite{adams11} employed greedy-point removal (GPR) and error-diffusion scheme together to achieve meshes of quality comparable to the original GPR scheme but at a much lower computational and memory complexities.  With the conjunction of smoothing operators, this method produces image triangulation of category I.  Adams also proposed a framework in \cite{adams13} for mesh generation by fixing various degrees of freedom available within that framework.  This method performs extremely well and produces meshes of higher quality than the GPR method, and is considered as a method of category I as well. By contrast, Li et al. \cite{li13} proposed a modified version of Rippa \cite{rippa92} and Garland-Heckbert (GH) \cite{garland95} frameworks which can generate single-layer mesh nodes on edges, and this framework generates triangulated images of category II. Another method of this category was proposed by Tu et al. \cite{tu13}, based on constrained Delaunay triangulations.  In this method, the approximating function is not required to be continuous everywhere but with discontinuities being permitted across constrained edges of triangles in triangulation.

Both categories of image triangulation generated by the methods mentioned above have their advantages and disadvantages. For the first category (double-layer vertices), the quality of image restoration is usually better because all vertices are well defined on images features thus the intensities of pixels during image restoration will not be affected by edges. As a result, the edges in the recovered images are sharp and the peak signal-to-noise ratio (PSNR) is usually larger.  While the restoration quality of methods in category I is high enough for subjective quality testing, the two layers must be very close to each other in order to have well-defined and sharp image edges. A consequence of this is that the resulting meshes always contain lots of thin and long triangles, which could cause large approximation errors when the meshes are to be used for numerical analysis (like finite element analysis). Additionally, in many applications, the direct communication between different materials should be maintained, meaning that no ``cushion'' layer between materials should be introduced in the meshes. Methods of category II avoid the small triangles and also the ``cushion'' layer problem, thus the mesh quality is usually better if proper steps are taken.  However, the vertices are defined on feature edges, where the nodal intensities are ambiguously defined. That is, the intensity of an edge point can be given by either side of the image edge. As will be shown in the experiments, the restored images often suffer from blurred and distorted feature edges if not properly addressed.

Because of the obvious limitations of methods in category I, we are more interested in a method that lies in the second category (single-layer approaches). However, in order to address the blurring and distortion problems often seen in existing approaches in this category, we propose a method based on the radial basis function (RBF) interpolation with the following improvements: 1) rather than considering only the Euclidean distances between vertices, our method also takes into consideration the image local orientations, yielding an anisotropic radial basis function (ARBF).  2) our method does not use intensities of vertices directly, but instead we utilize the intensities of triangles to eliminate the uncertainty of nodal intensities on feature edges.

The remainder of this paper is organized as the following.  Section \ref{subsec:mesh_generation} briefly summarizes our mesh generation method.  Section \ref{subsec:rbf_interpolation} introduces image restoration using traditional RBF interpolation.  The proposed image restoration is presented in Section \ref{subsec:arbf_interpolation}.  Section \ref{subsec:algorithm} shows the detailed algorithm of our proposed method.  Finally, Section \ref{sec:results} shows the experimental results and discussions.  Section \ref{sec:conclusion} concludes this paper.

\section{Methods}\label{sec:methods}
While mesh generation from images is not the main focus of the current paper, we will first give a brief summary of this step just for completion of the present work. The traditional (isotropic) radial basis function (RBF) interpolation is then introduced, followed by the proposed anisotropic RBF-based interpolation for image restoration from meshes. The detail of the implementation algorithm is given below as well.

\subsection{Adaptive Mesh Generation from Images}\label{subsec:mesh_generation}
A series of algorithms are used to generate high quality, feature-sensitive, and adaptive meshes from a given image. Firstly, three kinds of the sample points (namely, Canny's points, halftoning points, and uniform points) are generated. Secondly, a triangular mesh is generated from these points by using constrained Delaunay triangulation. The Canny's edge detector is employed to guarantee that important image features are preserved in the meshes. A halftoning-based sampling strategy is adopted to provide feature-sensitive and adaptive point distributions in the image domain. Finally, a Delaunay-triangulation is used to generate initial quality triangulation of the image. These steps are briefly summarized below.

\subsubsection{Canny Sample Points}\label{subsubsec:canny_sample_points}
Image edges are important features in an image and need to be preserved in the obtained meshes. Canny edge detector is a well-known method to deal with boundary extraction. In this paper, we use Canny edge detector to generate the initial Canny edge points and they are strictly attached to the boundary of the features of the image. However, the initial Canny edge points are too dense to yield quality meshes if all these edges are used as mesh nodes. In our method, we take the curvature information of every Canny's edge point into account and use the Principal Component Analysis (PCA) to determine the sampling density. The PCA method can detect the overall attribute of the neighbors of a certain size by a statistical way. After the PCA sampling, tiny features and features with high curvature have dense sample points and big features or features with straight lines have sparse sampled points.

\subsubsection{Halftoning Sample Points}\label{subsubsec:halftoning_sample_points}
The edge points generated by the Canny edge detector described above can only capture pixels on or near the image edges. In order to have a decent initial mesh, one has to scatter some more points in the non-edge regions of the image. We adopt the halftoning sample points based on the approach described in \cite{yang03}. This method generates the sample points using the second derivatives of an image, where most of the sample points are placed near the image features (edges).

\subsubsection{Uniform Sample Points}\label{subsubsec:uniform_sample_points}
Although the halftoning sample points can cover most non-edge regions of the image, it is possible that no point (either Canny or halftoning) is found in regions of almost constant intensities. We therefore generate some points uniformly to cover the rest of the images where the first two types of sample points are not located. A point (x, y) is said to be a valid uniform sample point if no Canny's or halftoning points are found in its neighborhood in a fixed distance.

\subsubsection{Mesh Generation via Constrained Delaunay Triangulation}\label{subsubsec:delaunay_triangulation}
The sample points found above are used to generate our triangular mesh for a given image by using the Delaunay triangulation. We employed a popular open source software {\em Triangle} \cite{shewchuk} for Delaunay triangulation. In order to guarantee the obtained meshes being well aligned with image edge features, we provide to {\em Triangle} with a set of line segments as additional constraints formed by connecting the Canny's sample points along the detected Canny's edges. With all the described strategies combined, we can generate high quality, feature-sensitive, and adaptive meshes from a given grayscale image. Some meshing examples will be shown in the result section below.

\subsection{Review of Radial Basis Function (RBF) Interpolation}\label{subsec:rbf_interpolation}
The traditional radial basis function interpolation is given by
\begin{equation}\label{eqn:rbf_interpolation_1}
f(\textbf{x})=\sum_{i=1}^N{w_i \phi(\|\textbf{x}-\textbf{x}_i\|)}
\end{equation}
where the interpolated function $f(\textbf{x})$ is represented as a weighted sum of $N$ radial basis functions $\phi(\cdot)$, each centered differently at $\textbf{x}_i$
and weighted by $w_i$.  Let $f_j=f(\textbf{x}_j)$.  By given conditions $f_j=\sum_{i=1}^N{w_i \phi(\|\textbf{x}_j-\textbf{x}_i\|)}$, the weights $w_i$ can be
solved by
\begin{equation}
\begin{bmatrix}
\phi_{11} & \cdots & \phi_{1N} \\
\vdots     & \ddots & \vdots \\
\phi_{N1} & \cdots & \phi_{NN}
\end{bmatrix}
\begin{bmatrix}
w_1 \\
\vdots \\
w_N
\end{bmatrix}
=
\begin{bmatrix}
f_1 \\
\vdots \\
f_N
\end{bmatrix}
\label{eqn:rbf_matrix}
\end{equation}
where $\phi_{ji}=\phi(\|\textbf{x}_j - \textbf{x}_i\|)$.  Once the unknown weights $w_i$ are solved, the image intensity at an arbitrary pixel can be calculated
by
\begin{equation}\label{eqn:rbf_interpolation_2}
s(\textbf{x})=\sum_{i=1}^N{w_i \phi(\|\textbf{x}-\textbf{x}_i\|)}
\end{equation}
In the traditional RBF method, the distance between the point $\textbf{x} \in \mathbb{R}^d $ and center $\textbf{x}_i \in \mathbb{R}^d$ is measured by Euclidean distance.  Let
$r=\|\textbf{x}-\textbf{x}_i\|$, commonly used radial basis functions include:
$$Gaussian: \phi(r)=e^{-(cr)^2}$$
$$Multiquadric\,\,(MQ): \phi(r)=\sqrt{r^2+c^2}$$
$$Inverse\,\,Multiquadric\,\,(IMQ): \phi(r)=\frac1{\sqrt{r^2+c^2}}$$
$$Thin\,\,Plate\,\,Spline\,\,(TPS): \phi(r)=r^2ln(r),$$
where $c$ is a shape parameter.  The shape parameter plays a major role in improving the accuracy of numerical solutions.  In general, the optimal
shape parameter depends on the densities, distributions and function values at the nodes.  However, it is difficult to assign different shape parameters
for each local domain.  Thus, choosing shape parameters has been an active topic in approximation theory \cite{wang02}.  Interested readers can refer to
\cite{vertnik09,kosec08,sarler06,vertnik06,divo07} for more details.

\begin{figure}[t]
\centering
\includegraphics[scale=0.5]{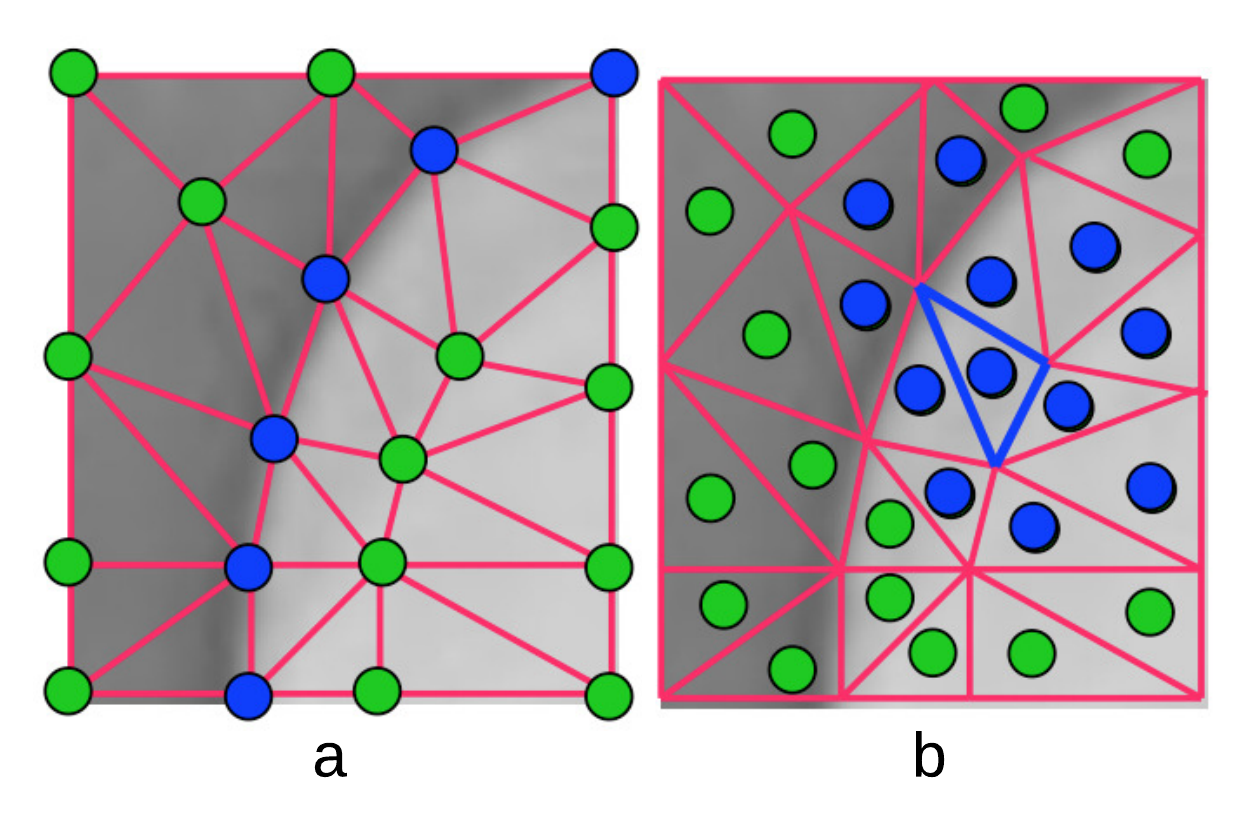}
\caption{Example of interpolation.  (a) Interpolation by vertices.  Green dots are vertices defined on feature.  Blue dots are vertices defined on feature
edge. (b) Interpolation by faces.  Green dots are face centers.  Blue dots are face centers used for interpolation of the intensities of pixels enclosed by the blue triangle.}
\label{fig:example_vertices_faces}
\end{figure}

One question about restoring image from triangular meshes is: what intensities should be used, intensities on vertices or intensities on faces? In the mesh generation approach described above, many vertices are located on image edges.  These vertices are good to capture image gradients (or orientations) but not for image intensities because there is an ambiguity in assigning intensity to a node defined on an edge, as illustrated for blue nodes in Fig. \ref{fig:example_vertices_faces} (a). Obviously, a very small change (or error) on the location of blues nodes would make a big interpolation difference if the mesh vertices are used as the nodal values in RBFs. A better way is to use face centers as the nodal values for RBF interpolations, which can eliminate the ambiguity and is less sensitive to mesh errors.  Fig. \ref{fig:example_vertices_faces} (b) shows this idea, where the face centers are more robust to the location changes of mesh vertices.  Results of vertex-based RBF interpolation and triangle-based RBF interpolation can be found in Fig. \ref{fig:results_1} (c) and \ref{fig:results_1} (d) in Section \ref{sec:results}.

Although using face centers performs better than the vertex-based RBF interpolation, the traditional RBF is isotropic in the sense that only the geometrical distance information is considered, which often causes blurring and distortion artifacts as can be seen in Fig. \ref{fig:results_1} (d). To capture the anisotropicity of the image features, the direction
of image edges has to be considered as well. Otherwise, nodes across feature edges may have strong influence on the pixel being interpolated.  Fig. \ref{fig:interpolation_rbf_arbf} (a) shows the cause of the blurred edge problem.  $\textbf{x}$ is the pixel whose intensity we want to
find out. The intensities on nodes $\textbf{x}_1$ and $\textbf{x}_2$ are two of the neighbors used for interpolation.  The weights of them are determined only by the Euclidean
distance to $\textbf{x}$ based on the definition of traditional RBF.  However, $\textbf{x}_1$ is on the other side of the feature edge, so it should have much less
influence on $\textbf{x}$ than $\textbf{x}_2$.  The isotropic RBF has a hyper-spherical support domain which cannot satisfy this data-dependent requirement.  Thus the
intensity on $\textbf{x}$ is blurred by $\textbf{x}_1$.  As a contrast, Fig. \ref{fig:interpolation_rbf_arbf} (b) shows the anisotropic RBF (ARBF) interpolation.
The support domain of ARBF is a hyper-ellipsoid.  By choosing proper shape parameter, the support domain could rule out the interfering node $\textbf{x}_1$,
or give insignificant weight to node $\textbf{x}_1$.  Thus the blurring effect will be eliminated and sharp features can be well retained. In the following subsections we will elaborate on the detail of designing anisotropic radial basis functions for image restoration.

\subsection{Anisotropic Radial Basis Function (ARBF) interpolation}\label{subsec:arbf_interpolation}
The main difference between the isotropic and anisotropic RBFs is the definition of distance metrics used. As in \cite{casciola10}, the anisotropic RBF is defined as follows:

\begin{figure}[t]
\centering
\includegraphics[scale=0.6]{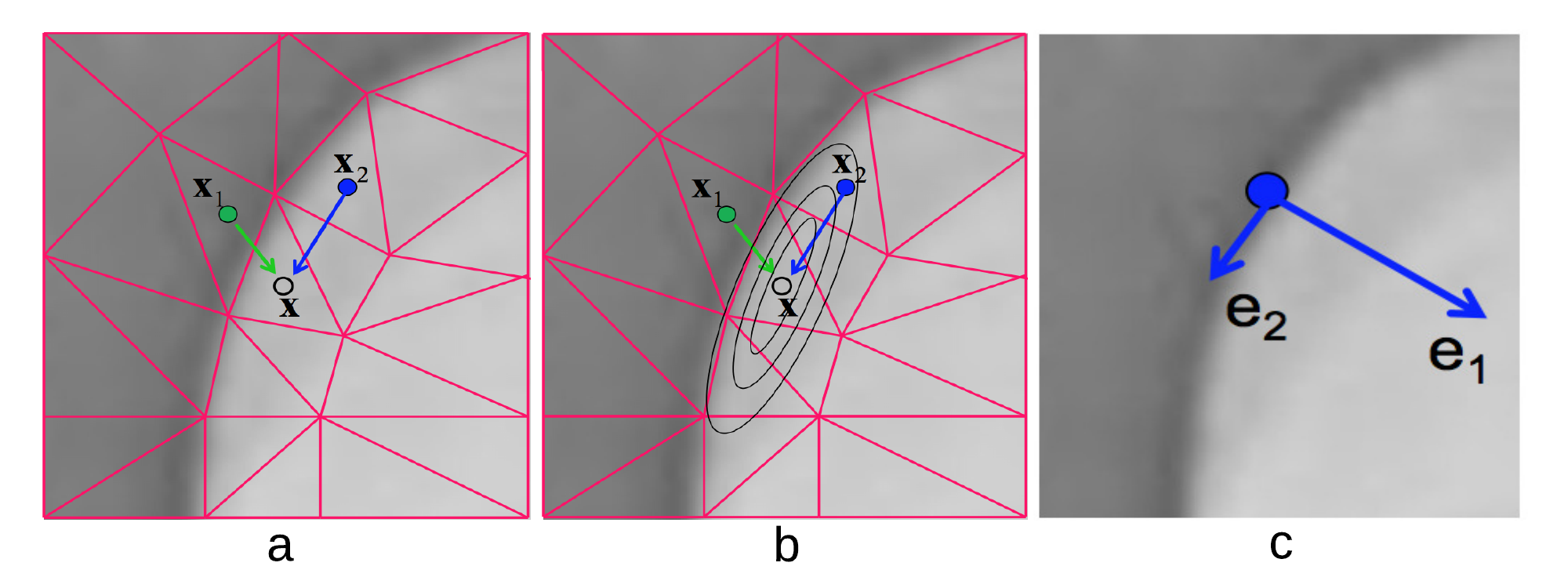}
\caption{Interpolation schemes.  (a) Isotropic RBF interpolation.  (b) Anisotropic RBF interpolation.  (c) Eigenvectors on an edge pixel.
$\textbf{e}_1$ shows the normal direction.  $\textbf{e}_2$ shows the tangent direction.}\label{fig:interpolation_rbf_arbf}
\end{figure}

\begin{definition}
Given $N$ distinct points $X=\{\textbf{x}_j \in \mathbb{R}^d\}_{j=1,\ldots,N}$ and a $d \times d$ positive definite matrix $\textbf{T}$,
the anisotropic radial basis function associated with a radial basis function $\Phi_j(\cdot)=\phi(\|\cdot-\textbf{x}_j\|_2)$ is defined by
\begin{equation}
\Phi_{\textbf{T},j}(\cdot):=\phi(\|\cdot-\textbf{x}_j\|_\textbf{T}),
\end{equation}
where $\|\textbf{x}\|_\textbf{T}=\textbf{x}^T\textbf{T}\textbf{x}$.
\end{definition}

The support domain of ARBF is hyper-ellipsoid instead of a hyper-sphere in traditional RBF.
Its center is $\textbf{x}_j$, associated with the quadratic form $(\textbf{x}-\textbf{x}_j)^T(\textbf{x}-\textbf{x}_j)$.
Interested readers can refer to \cite{casciola06} and \cite{casciola07} for more details of ARBF.

To construct the metric $\textbf{T}$, , we use the image structure tensor
$$G_\sigma * \begin{bmatrix}I_x^2 & I_xI_y \\
					     I_xI_y & I_y^2\end{bmatrix}$$
where $G_\sigma$ is the Gaussian smooth operator, and $\begin{bmatrix}I_x \\ I_y\end{bmatrix}$ is the image gradient at a pixel.  Two
eigenvectors $\textbf{e}_1$ and $\textbf{e}_2$ are the normal and tangent directions of the edge, respectively, as shown in Fig. \ref{fig:interpolation_rbf_arbf} (c).
The corresponding eigenvalues are $\lambda_1$ and $\lambda_2$.  The anisotropic metric is defined by
\begin{equation}\label{eqn:metric_T}
\textbf{T}= \begin{bmatrix}\textbf{e}_1 \textbf{e}_2\end{bmatrix}
\begin{bmatrix}\lambda_1 & 0 \\ 0 & \lambda_2\end{bmatrix}
\begin{bmatrix}\textbf{e}_1^T \\ \textbf{e}_2^T\end{bmatrix}
\end{equation}
Similar to the isotropic RBF but with a modified distance metric, the ARBF image interpolation problem becomes
\begin{equation}\label{eqn:arbf_interpolation}
s'(\textbf{x})=\sum_{i=1}^N{w'_i \phi(\|\textbf{x}-\textbf{x}_i\|_\textbf{T})}
\end{equation}

Please note that the matrix in equation (\ref{eqn:rbf_matrix}) should also be updated accordingly with the new distance metric $\textbf{T}$. Therefore, the new set of weights $w'_i$ would be different from the weights $w_i$ in the isotropic RBF interpolation.

\subsection{Algorithms}\label{subsec:algorithm}
The following algorithm shows the steps of the proposed approach for image restoration from triangular meshes.  The major step is the ARBF interpolation which comprises of two sub-steps.  First, the weight coefficients are solved by using the new distance metric $\textbf{T}$.  As stated in section \ref{subsec:rbf_interpolation}, this is done by taking intensities at triangle centers.  Then the weights are applied to equation (\ref{eqn:arbf_interpolation}) to restore the intensity at each pixel.
\\
\\
\noindent
{Algorithm: Image Reconstruction}
\label{alg:arbf_interpolation}
\begin{verbatim}
ImageReconstruction()
{
    loadMesh();
    calculateTriangleCenters();
    findNeighbors();
    calculateEigenvalues();
    computeMetrics();
    ARBFInterpolation();
    printResult();
}

ARBFInterpolation()
{
    for (every triangle centers)
        solveCoefficients();
    end
	
    for (every triangles)
        for (every pixel in current triangle)
            applyCoefficientsToInterpolation();
        end
    end
}
\end{verbatim}

\section{Results and Discussion}\label{sec:results}
Numerous experiments have been conducted on publicly available images by using the proposed approaches and the image restoration results are all promising. Due to the space limit, we will only consider the well-known ``Lena'' image and three medical images of different sizes.

\begin{figure}[t]
	\centering
	\includegraphics[scale=0.43]{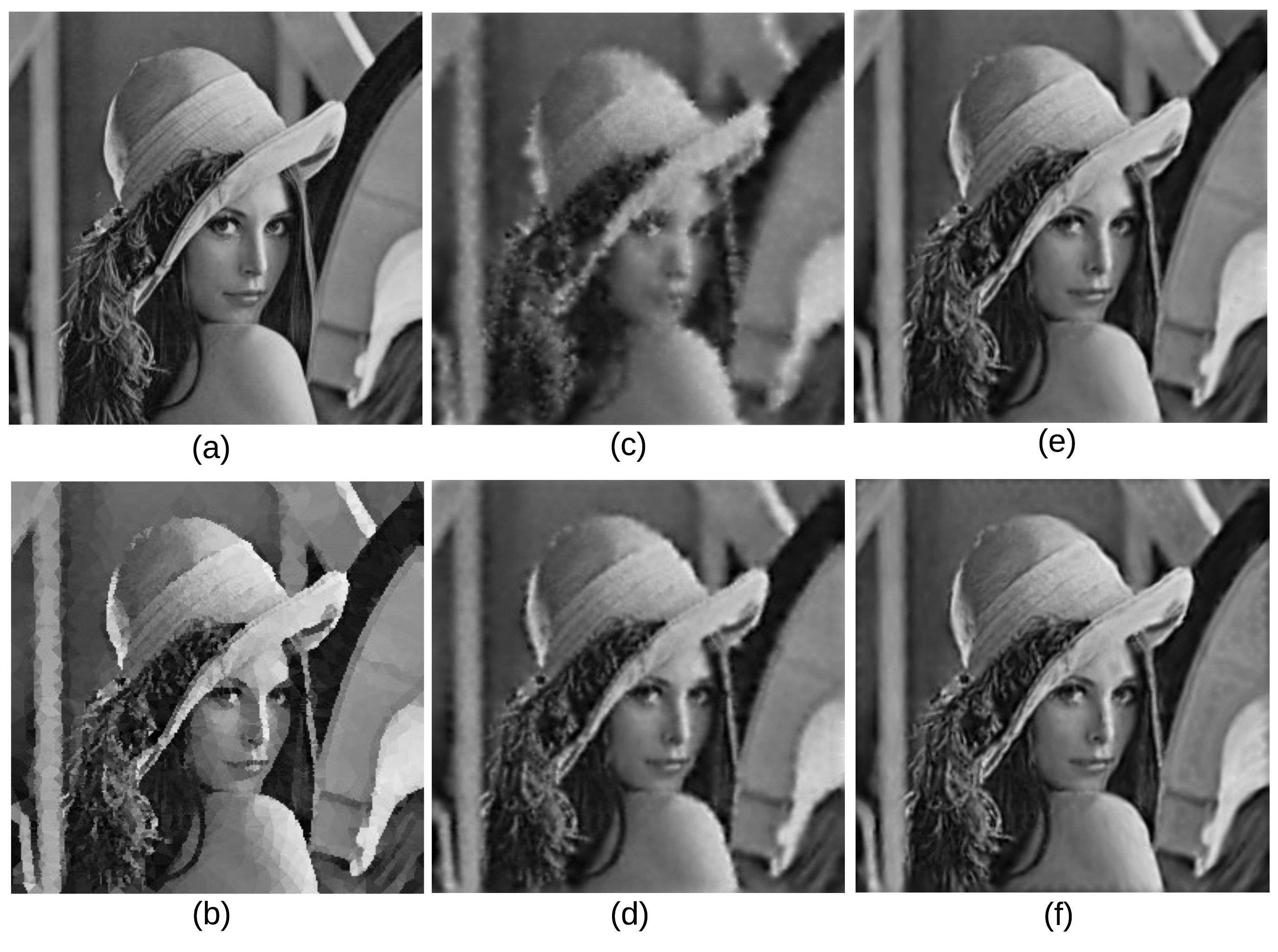}
	\caption{Summary of restoration of Lena.  (a) Original Lena image.  (b) Result of piecewise interpolation.  (c) Result of vertex-based iso-RBF interpolation.  (d) Result of triangle-based iso-RBF interpolation.  (e) Result of triangle-based ARBF interpolation using MQ basis.  (f) Result of triangle-based ARBF interpolation using IMQ basis.}\label{fig:results_1}
\end{figure}

Fig. \ref{fig:results_1} (a) is the original Lena image of size $256 \times 256$ pixels.  Fig. \ref{fig:results_1} (b) is the result of assigning a constant intensity to all pixels in a mesh triangle (so-called piecewise interpolation).  As we can see, this result shows heavy mosaic effect.  Fig. \ref{fig:results_1} (c) is the result of iso-RBF interpolation using intensities on vertices.  As previously stated on section \ref{subsec:arbf_interpolation}, the ambiguity of intensities on vertices blurred the result.  Fig. \ref{fig:results_1} (d) is the result of iso-RBF interpolation using intensities on triangle centers.
In this case, there is no ambiguity of intensities.  So the result is much better comparing to Fig. \ref{fig:results_1} (c).  However, the feature edges are still blurred and some distortions are clearly seen because of the lack of directional information used in isotropic RBF.  Fig. \ref{fig:results_1} (e) is the result of ARBF interpolation using intensities on triangle centers with multi-quadrics (MQ) basis function.  The result is much better thanks to a modified distance metric that incorporates both geometric distances and data-dependent feature orientations.  Fig. \ref{fig:results_1} (f) is similar to Fig. \ref{fig:results_1} (e), except that the basis function is inverse multi-quadrics (IMQ).  We have also tested other basis functions like Gaussian and Thin-Plate-Spline (TPS).  However, it is hard to find a proper shape parameter to get a reasonable result for Gaussian, and the TPS interpolation doesn't converge.

\begin{figure}
	\centering
	\includegraphics[scale=0.46]{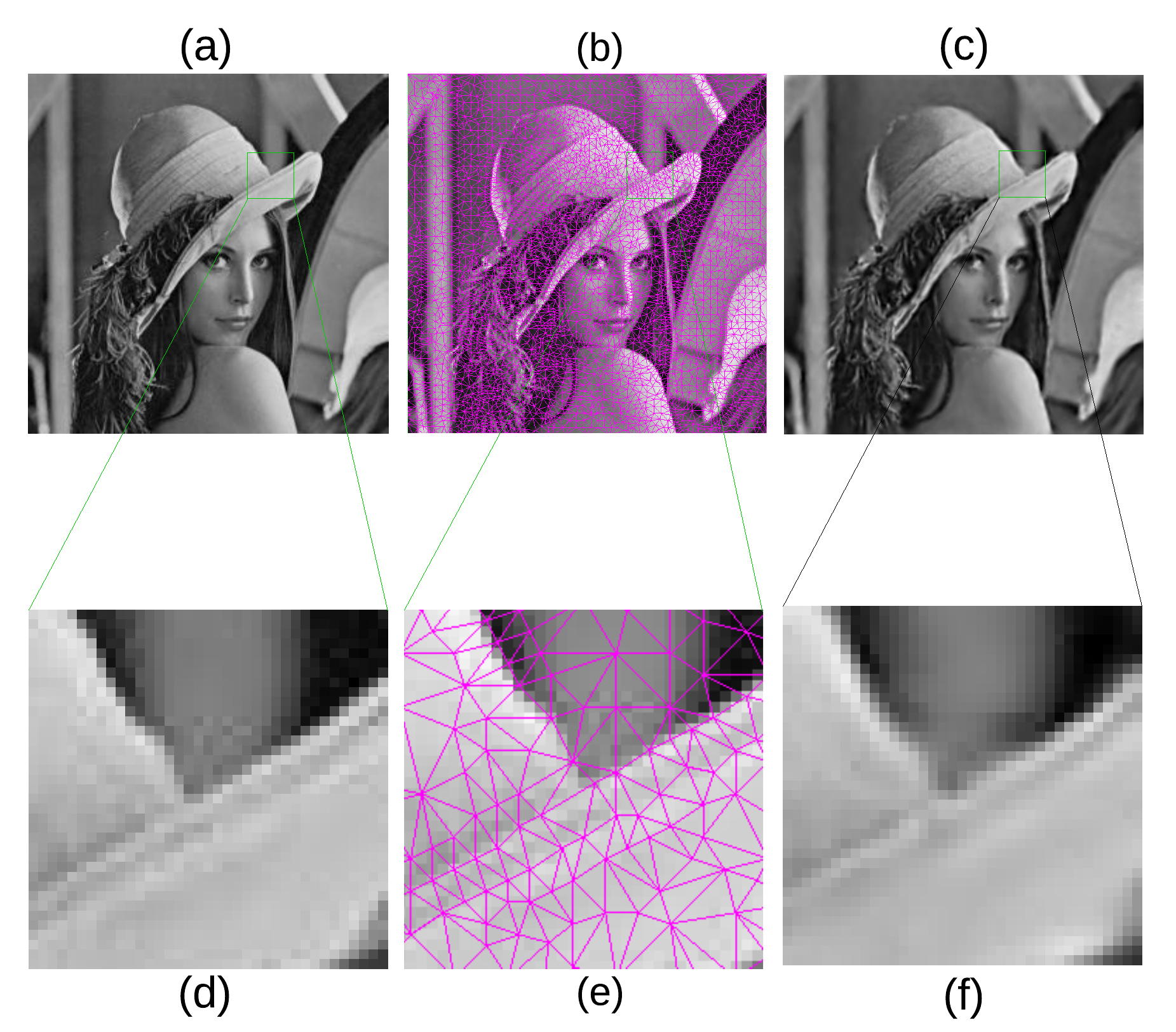}
	\caption{Details of Lena.  (a) Original Lena image.  (b) Generated mesh of (a).  (c) Result of triangle-based ARBF interpolation using MQ basis.  (d)--(f) are zoomed-in views of (a)--(c), respectively.}\label{fig:details_lena}
\end{figure}

\begin{figure}
	\centering
	\includegraphics[scale=0.46]{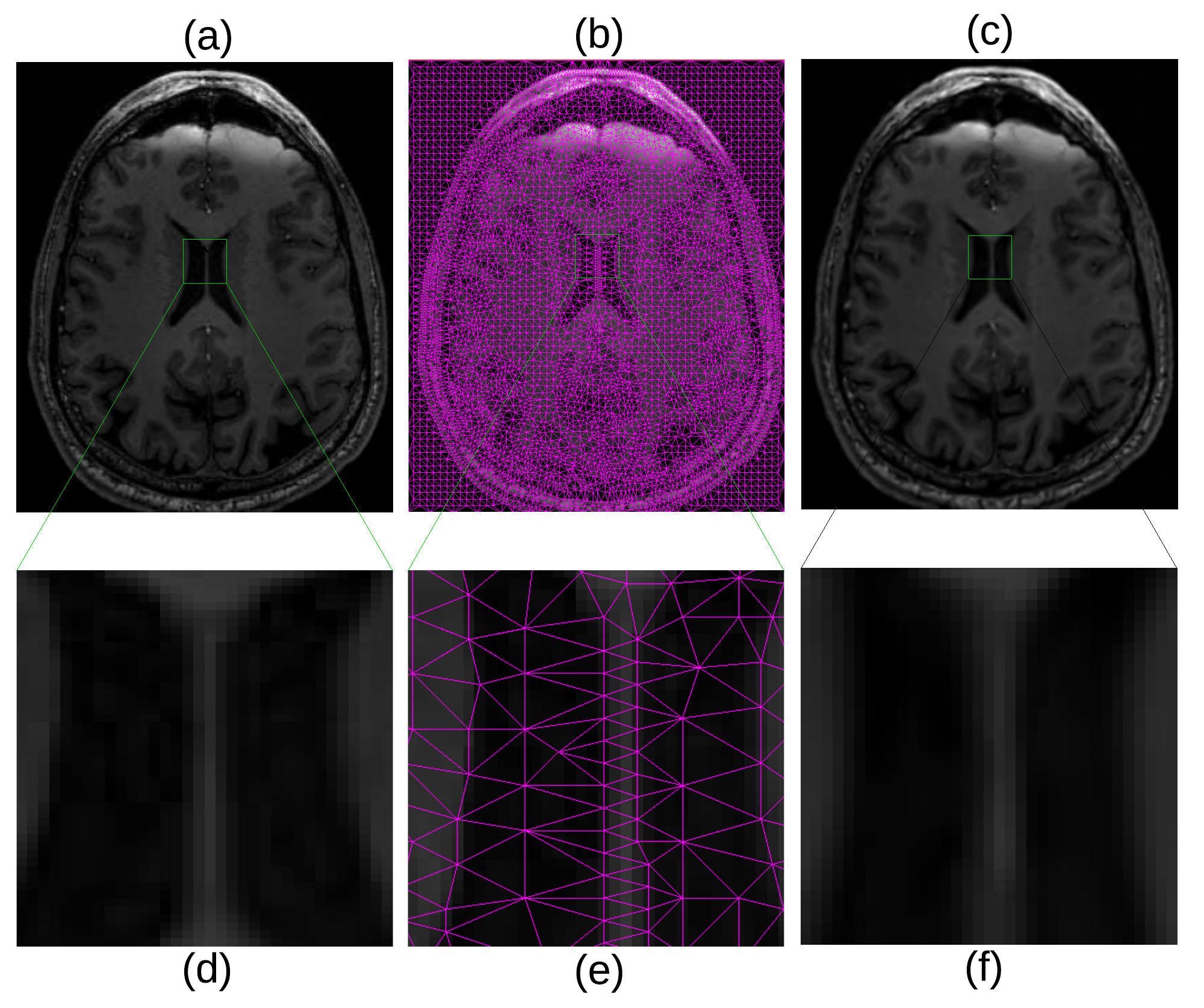}
	\caption{Details of brain MRI.  (a) Original brain MRI.  (b) Generated mesh of (a).  (c) Result of triangle-based ARBF interpolation using MQ basis.  (d)--(f) are zoomed-in views of (a)--(c), respectively.}\label{fig:details_brain}
\end{figure}

\begin{figure}
	\centering
	\includegraphics[scale=0.46]{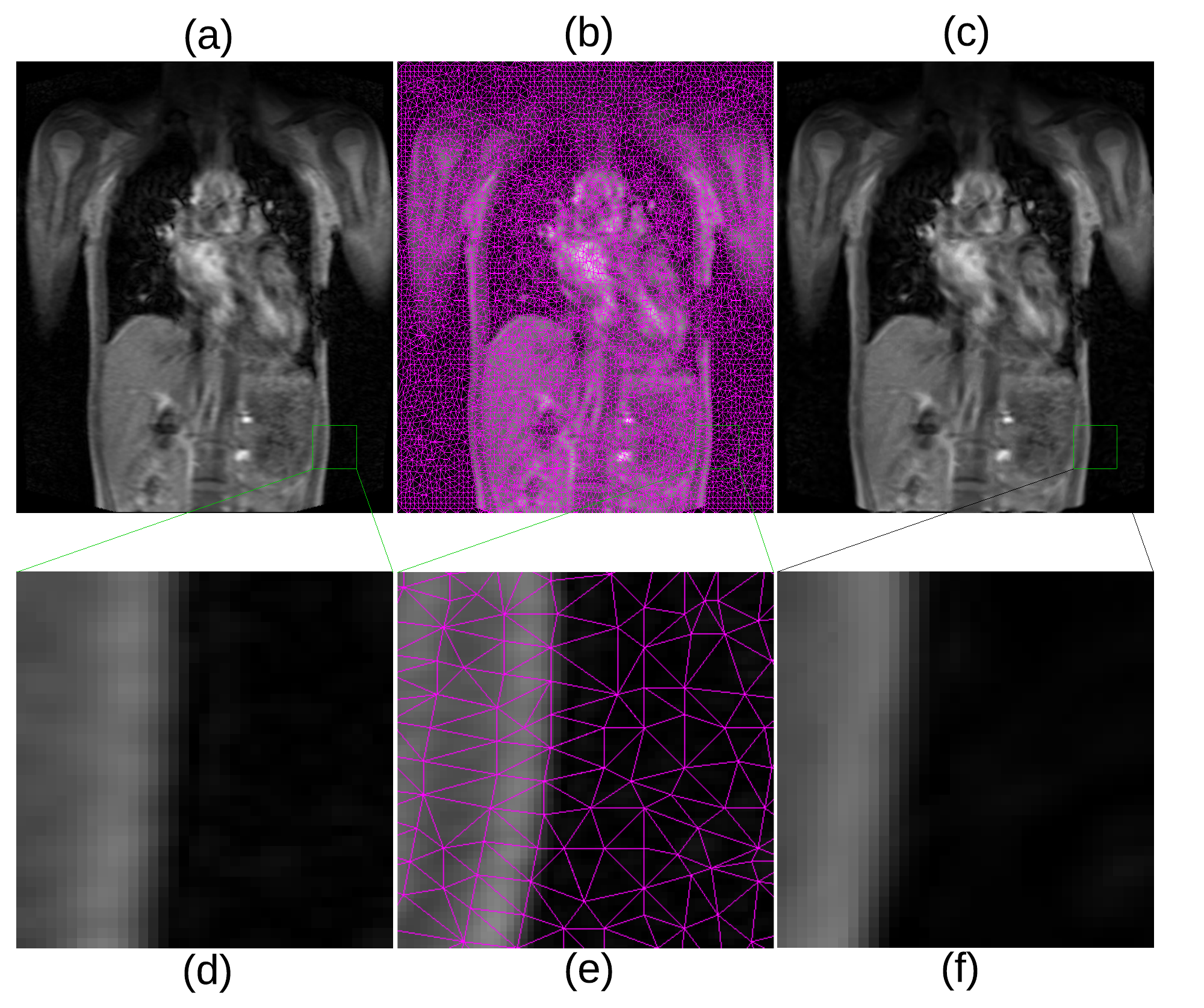}
	\caption{Details of breast MRI.  (a) Original breast MRI.  (b) Generated mesh of (a).  (c) Result of triangle-based ARBF interpolation using MQ basis.  (d)--(f) are zoomed-in views of (a)--(c), respectively.}\label{fig:details_breast}
\end{figure}

\begin{figure}
	\centering
	\includegraphics[scale=0.46]{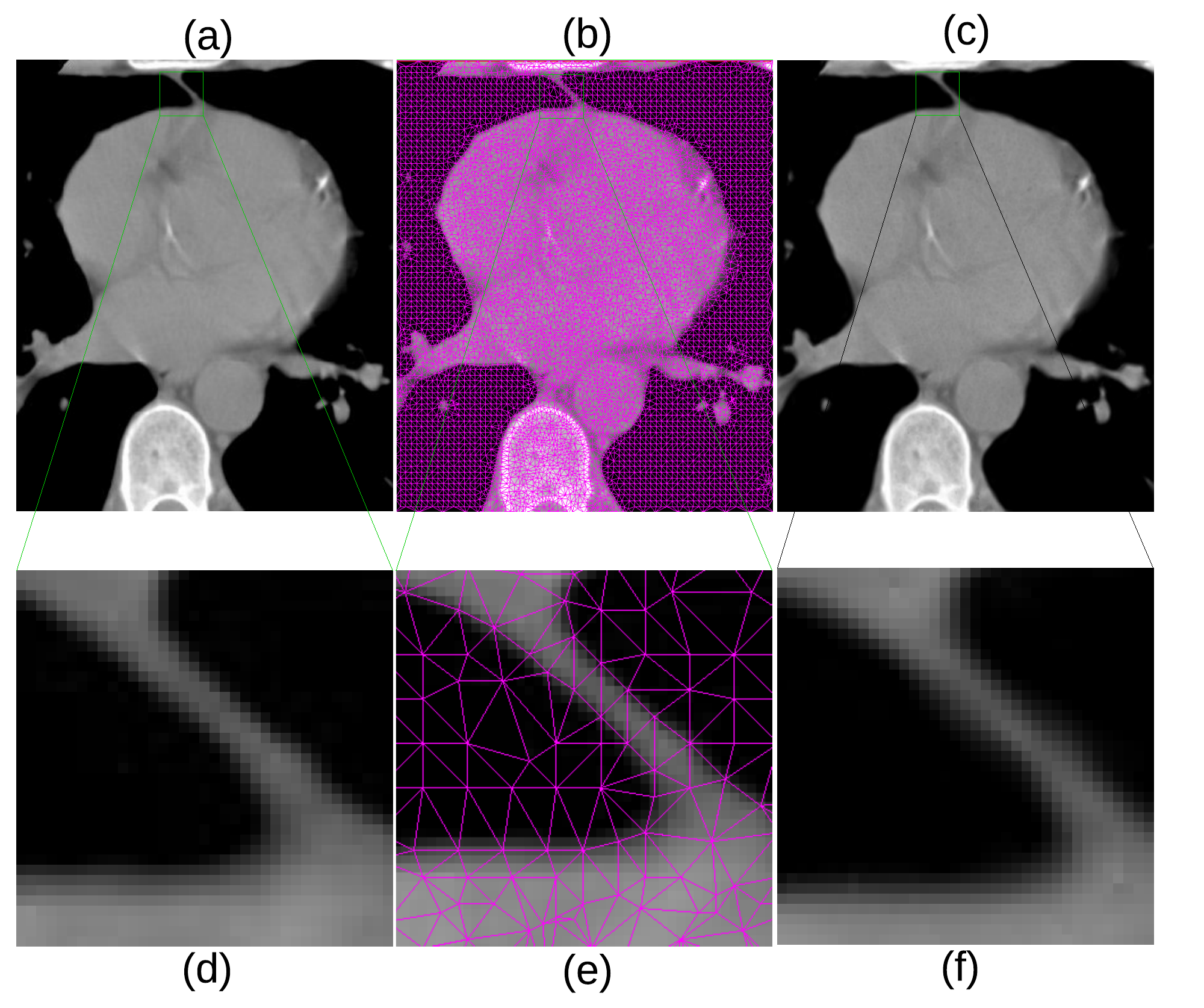}
	\caption{Details of CT-scanned image of heart.  (a) Original heart image.  (b) Generated mesh of (a).  (c) Result of triangle-based ARBF interpolation using MQ basis.  (d)--(f) are zoomed-in views of (a)--(c), respectively.}\label{fig:details_heart}
\end{figure}

In Fig. \ref{fig:details_lena}, more details of the Lena experiment are shown. (a) is the original Lena image, the same as Fig. \ref{fig:results_1} (a).  (b) is the mesh generated by the method outlined in section \ref{subsec:mesh_generation}.  (c) is the recovered image, which is the same as Fig. \ref{fig:results_1} (e). To visually see the generated mesh and compare the difference between the original and restored images, Fig. \ref{fig:details_lena} (d)--(f) are the zoomed-in views of (a)--(c), respectively.  As the results show, the mesh quality is high enough for subsequent numerical analysis and the the recovered image is very close to the original one. As a matter of fact, the restored image looks smoother due to the smooth radial basis functions used, and the sharp edge features are well preserved.  Fig. \ref{fig:details_brain} shows the original brain MRI, its generated mesh, and the result of ARBF interpolation using intensities on triangle centers with the MQ basis function.  The zoomed-in views show the quality of mesh and restoration as well.  Fig. \ref{fig:details_breast} shows another MRI experiment of breast.  Fig. \ref{fig:details_heart} shows a CT-scanning experiment. From all these examples, one can see the effectiveness of the proposed approaches for image mesh generation and feature-preserving restoration.

To give a quantitative evaluation of the restored images, we use the widely-used peak signal-to-noise ratio (PSNR) as defined below:
$$
PSNR=20*\log_{10}\left(\frac{255}{RMSE}\right),
$$
where
$$
RMSE=\sqrt{\frac{\sum_{i=0}^{M-1}\sum_{j=0}^{N-1}{\left[O(i,j)-I(i,j)\right]^2}}{M * N}},
$$
where $M$ and $N$ are the dimensions of the image. $O(i,j)$ is the original intensity at pixel $(i,j)$ and $I(i,j)$ is the interpolated intensity at $(i,j)$.

Table \ref{tab:summary1} gives a summary of the Lena image using different restoration approaches. The compression ratio in the table means the ratio of the number of vertices in the mesh vs. the number of pixels in the original image. As we can see, the restored image with the anisotropic RBF interpolation gives the best PSNR score. Table \ref{tab:summary2} summarizes the other three data sets, where the running time of image restoration for each case is included and measured on a PC with 1.8 GHz CPU and 2 GB RAM. The proposed algorithms were implemented in C programming and will be released to the public.

\section{Conclusions}\label{sec:conclusion}
The present paper describes a nonlinear interpolation method by using anisotropic radial basis functions and structure tensor driven metrics.  Using the proposed methods, an original image can be stored and processed in the mesh format with some nice advantages including less storage requirement, faster transmission speed, and more efficient image processing due to the significantly reduced number of mesh nodes as opposed to the number of pixels in the original image.  The generated meshes, after some post-processing such as mesh-based segmentation, can be readily used for further numerical analysis. The present image restoration algorithm provides an effective way to restore the image with an arbitrary super-resolution from a mesh representation, serving as a decoding algorithm for the mesh-based image coding technique. The anisotropic RBF algorithm can be used as a de-blurring process as well with sharp features well preserved in the images.

As the image restoration algorithm shows, the time complexity of the function ARBFInterpolation() is $O(m \times n)$, where $m$ is the number of triangles and $n$ is the number of pixels inside of a triangle. In case of 3D images or very large 2D images, the running time could be very expensive. One of our further investigations would be the parallel implementation of the proposed algorithm using GPU programming. Fortunately the present method is very straightforward to parallelize to accelerate the computations.  Additionally we are also interested in the mesh-based image segmentation by using the adaptive meshes generated from the original images, and in applying the segmented meshes to image-based numerical analysis.

\begin{table}
	\centering
	\caption{Summary of the Lena image (Fig. \ref{fig:results_1}).}\label{tab:summary1}
	\begin{tabular}[b]{l|c|c}
		\hline
		Lena (size is $256 \times 256$, compression ratio is 6\%) & PSNR (db) & Shape Parameter \\
		\hline
		Piecewise Interpolation & 22.9703 & 0.5 \\
		Triangle-based ISO-RBF Interpolation & 26.7367 & 0.5 \\
		Triangle-based ARBF Interpolation (MQ) & 28.2088 & 0.5 \\
		Triangle-based ARBF Interpolation (IMQ) & 27.1836 & 1.8 \\
		\hline
	\end{tabular}
\end{table}

\begin{table}
	\centering
	\caption{Summary of the three medical images (Fig. \ref{fig:details_brain}--\ref{fig:details_heart}).}\label{tab:summary2}
	\begin{tabular}[b]{l|c|c|c|c|c}
		\hline
		Data & Size & Compression Ratio & PSNR (db) & Shape Parameter & Time (seconds) \\
		\hline
		Brain & $285 \times 341$ & 6\% & 15.7058 & 0.5 & 1.71 \\
		Breast & $512 \times 512$ & 5\% & 11.8763 & 0.5 & 8.70 \\
		Heart & $356 \times 396$ & 5\% & 10.5208 & 0.5 & 2.73 \\
		\hline
	\end{tabular}
\end{table}

\bibliographystyle{splncs}
\bibliography{image_triangulationbib}

\end{document}